\documentclass[preprint,12pt,3p]{elsarticle}

\usepackage{graphicx}
\usepackage{amsmath}

\usepackage{amssymb}

\usepackage{hyperref}
\hypersetup{
    colorlinks=true,
    linkcolor=blue,
    filecolor=magenta,      
    urlcolor=cyan,
    pdftitle={Overleaf Example},
    pdfpagemode=FullScreen,
    }

\journal{arXiv}

\begin{document}

\begin{frontmatter}

\title{A Multi-Task Deep Learning Framework for Skin Lesion Classification, ABCDE Feature Quantification, and Evolution Simulation}

\author[1]{Harsha Kotla}
\author[1]{Arun Kumar Rajasekaran}
\author[2]{Hannah Rana}

\address[1]{Department of Chemistry,
University of Cambridge, Cambridge, United Kingdom}
\address[2]{Harvard Ophthalmology AI Lab, Harvard Medical School, Boston, MA, USA}

\date{} 

\begin{abstract}
Early detection of melanoma has grown to be essential because it significantly improves survival rates, but automated analysis of skin lesions still remains challenging. ABCDE, which stands for Asymmetry, Border irregularity, Color variation, Diameter, and Evolving, is a well-known classification method for skin lesions, but most deep learning mechanisms treat it as a black box, as most of the human interpretable features are not explained. In this work, we propose a deep learning framework that both classifies skin lesions into categories and also quantifies scores for each ABCD feature. It simulates the evolution of these features over time in order to represent the E aspect, opening more windows for future exploration. The A, B, C, and D values are quantified particularly within this work. Moreover, this framework also visualizes ABCD feature trajectories in latent space as skin lesions evolve from benign nevuses to malignant melanoma. The experiments are conducted using the HAM10000 dataset that contains around ten thousand images of skin lesions of varying stages. In summary, the classification worked with an accuracy of around 89 percent, with melanoma AUC being 0.96, while the feature evaluation performed well in predicting asymmetry, color variation, and diameter, though border irregularity remains more difficult to model.  Overall, this work provides a deep learning framework that will allow doctors to link ML diagnoses to clinically relevant criteria, thus improving our understanding of skin cancer progression. 
\end{abstract}


\end{frontmatter}



\section{Introduction}
\label{introduction}

Melanoma, an aggressive form of skin cancer, is one of the leading causes of death due to skin cancer \cite{Jutte2024Integrating}. Early diagnosis is important because the 5-year survival rate exceeds 90\% for early-stage melanoma, but drops below 20\% for advanced stages \cite{Jutte2024Integrating}. In order to differentiate between harmful and harmless lesions, dermatologists utilize the ABCDE method. “A” stands for “asymmetry,” as malignant skin lesions often appear to be uneven; “B” stands for “border irregularity,” as scientists search for jagged or notched edges; “C” stands for “color variation”; “D” stands for diameter, as larger lesions are more likely to be malignant; and “E” stands for “evolving,” as skin lesions evolve over time \cite{Veeramani2024DDCNNF}. If a lesion displays two or more of the attributes described above, the lesion is most likely harmful melanoma. The ABCDE criteria are effective because they are easy to understand and to screen for suspicious lesions \cite{Veeramani2024DDCNNF}. 

Many deep learning techniques are proficient at classifying skin lesion images as either benign, malignant, or one of several other clinically recognized categories found in datasets such as HAM10000 \cite{Abdullah2024Intelligent} \cite{Tschandl2018HAM10000}. Convolutional neural networks are utilized most commonly, as they are trained to classify images. However, these CNN models lack clear interpretability because they only provide point predictions of lesion type without explanation. In the context of melanoma detection, this lack of explanation  can hinder clear, interpretable diagnoses. 

The “E” feature of ABCDE has also proven challenging to evaluate via deep learning methods. Single static images fail to capture such a change. Some mobile applications ask users to input their moles periodically to observe any changes, but this is difficult to enforce. However,  using AI to predict and visualize how a lesion changes over time is a promising direction. Recent advances in medical imaging have made it possible to create realistic transformations of medical images. For example, Jütte et al. (2024) utilized a CycleGAN to create a sequence of dermoscopic images that show the potential of a benign nevus transforming into a malignant melanoma \cite{Jutte2024Integrating}. As discussed above, the quantification of ABCDE features changing over time as well as the actual images changing can improve our understanding about melanoma growth patterns \cite{Jutte2024Integrating}. 

In this work, we propose a deep-learning framework that combines classification, ABCDE feature quantification, and feature evolution simulation. First, we design a CNN that learns to predict both the lesion’s class and quantify continuous values representing each ABCDE criterion. Second, we develop a strategy for obtaining quantitative ABCDE labels from dermoscopic images by image processing and expert knowledge. This enables supervised training of the feature regression branch. Subsequently, we introduce a module to simulate the temporal evolution of lesions. Generative adversarial networks and sequential interpolation are used so that this system can produce a plausible future state of a given lesion and track how the ABCDE scores change. 

Automated ABCDE analysis was already prevalent before the era of deep learning. Researchers wanted to mimic the ABCD rule of dermoscopy with computer algorithms. Early systems computed hand-crafted image features corresponding to asymmetry, border irregularity, color variegation, and lesion diameter. For example, in 2001, Ganster et al. extracted 122 features related to the ABCD criteria and applied conventional classifiers for automated melanoma recognition \cite{Ganster2001Automated}. Asymmetry was quantified via shape moments, border irregularity via fractal dimension or edge abruptness, color variegation via histogram analysis, and diameter via lesion area in pixels \cite{Veeramani2024DDCNNF}. These pipelines showed that it is feasible to use computer vision for melanoma screening, but they often need expert parameter tuning and struggle with variations in image quality. A recent survey by Celebi et al. (2022) summarizes many such efforts to automate parts of the ABCD rule \cite{Celebi2007Methodological}. In general, while these approaches brought interpretability (each feature could be reported), their accuracy was typically lower than that of data-driven deep learning, which can learn more complex representations. 

In addition, the success of CNNs in image recognition has led to numerous applications in dermatology. CNN models have become very accurate while classifying lesions into categories such as melanoma, basal cell carcinoma, nevus, etc. For example, on the HAM10000 dataset, approaches like an EfficientNet or an ensemble of CNNs reach high overall accuracy (often 85–90\%+) in 7-class classification \cite{Veeramani2024DDCNNF}. Some research focuses on binary classification (melanoma vs. benign), as they report very high specificity and good sensitivity \cite{Veeramani2024DDCNNF}. Still, as described above, these models lack interpretability. Other researchers have used saliency maps and class activation maps in order to improve interpretability. However, these pixel-level explanations do not directly communicate which clinical features are most explanatory of the diagnosis. This motivates the usage of the ABCDE criteria explicitly. Choi et al. (2024) proposed an “ABC ensemble model” that preprocesses images to emphasize asymmetry, border, and color regions as they feed them into specialized network branches; these are then combined for classification \cite{Choi2024ABCEnsemble}. Still, their model did not output quantitative values for each criterion, even though they were able to use ABCD rule knowledge to improve classification performance. Notably, they omitted the diameter feature because the scaling in the data was inconsistent \cite{Choi2024ABCEnsemble}. 

In summary, this work aims to build upon prior research by combining the interpretability of rule-based features with the accuracy of deep learning. This work also extends on previous research by adding the ability to simulate temporal changes. 

\section{Methodology}
\label{methodology}
\subsection{Overview of the Framework}
The framework contains two main components: a CNN to perform lesion classification and ABCDE feature regression from a dermoscopic image, and also an evolution simulation module that shows how ABCDE features might progress over time. Given a dermoscopic image of a skin lesion, we first optionally preprocess it (including lesion segmentation and color normalization). The multi-task CNN then processes the image to output both a class prediction and a set of numeric scores corresponding to A, B, C, and D features. The CNN is optimized by using a combined loss that includes classification error and regression error on the ABCDE scores. After this model is trained, it can provide an interpretation for its diagnosis by showing the ABCDE scores. For the evolution simulation, we take a lesion image and generate a sequence of future images showing increasing malignancy. This CNN model is applied to each generated frame to track how the ABCDE scores change; it basically gives a trajectory of the features. Additionally, the model’s internal representation is used to predict feature values without image generation.

 \subsection{Multi-Task CNN Architecture}

This multi-task deep learning model is built based on a convolutional neural network that first extracts a shared representation of the input image, followed by two “heads” (output branches). These branches consist of one for lesion classification and one for ABCDE feature regression. This design allows the model to learn common visual features that are useful for both tasks \cite{Kawahara2019SevenPointChecklist}. Still, the separate heads specialize in their respective outputs. We experimented with several backbone architectures like ResNet50 and EfficientNet, but we ultimately chose ResNet50 due to its balance of depth and efficiency \cite{He2016ResNet} \cite{Tan2019EfficientNet}. The classification head is a dense layer that produces a probability distribution over the lesion classes. The regression head is a fully-connected dense layer to produce 4 values corresponding to [A, B, C, D] (E is not supervised). Overall, we use linear outputs with appropriate activation/normalization; we do this to make sure that the feature values fall in a reasonable range (0 to 1 for A, B, C and a scaled range for D as discussed later). 

The input for this network is a dermoscopic image. The images are rescaled to 224 by 224 pixels, and we also normalize all the color channels. The ResNet50 backbone processes the image through a series of convolutional layers. This gives a final feature map which is global-average-pooled to a 2048- dimensional feature vector. This vector represents high-level information about the lesion. Also, the network is trained to predict the ABCDE features, so the vector encodes information relevant to asymmetry, border, color, and others, in addition to other features useful to classify lesions. 

After that, we add a fully connected layer called the classification head. This takes in the 2048 feature vector and produces logits for each of the seven classes for HAM10000 \cite{Tschandl2018HAM10000}. These include nv, mel, bcc, akiec, bkl, df, and vasc and correspond to melanocytic nevus, melanoma, basal cell carcinoma, actinic keratosis, benign keratosis, dermatofibroma, and vascular lesion \cite{Mader2018SkinCancerMNIST}. During training, we use a cross-entropy loss for this head.

The regression head  maps the same feature vector to five numeric outputs representing [A, B, C, D, E]. No activation (linear output) is applied for regression. However, these values are constrained through the training data scaling and loss function; this is so that the outputs remain in plausible ranges. Mean squared error loss is used here as it sums over the 5 features for this head. 

\begin{figure}[h]
    \centering
    \includegraphics[width=0.65\linewidth]{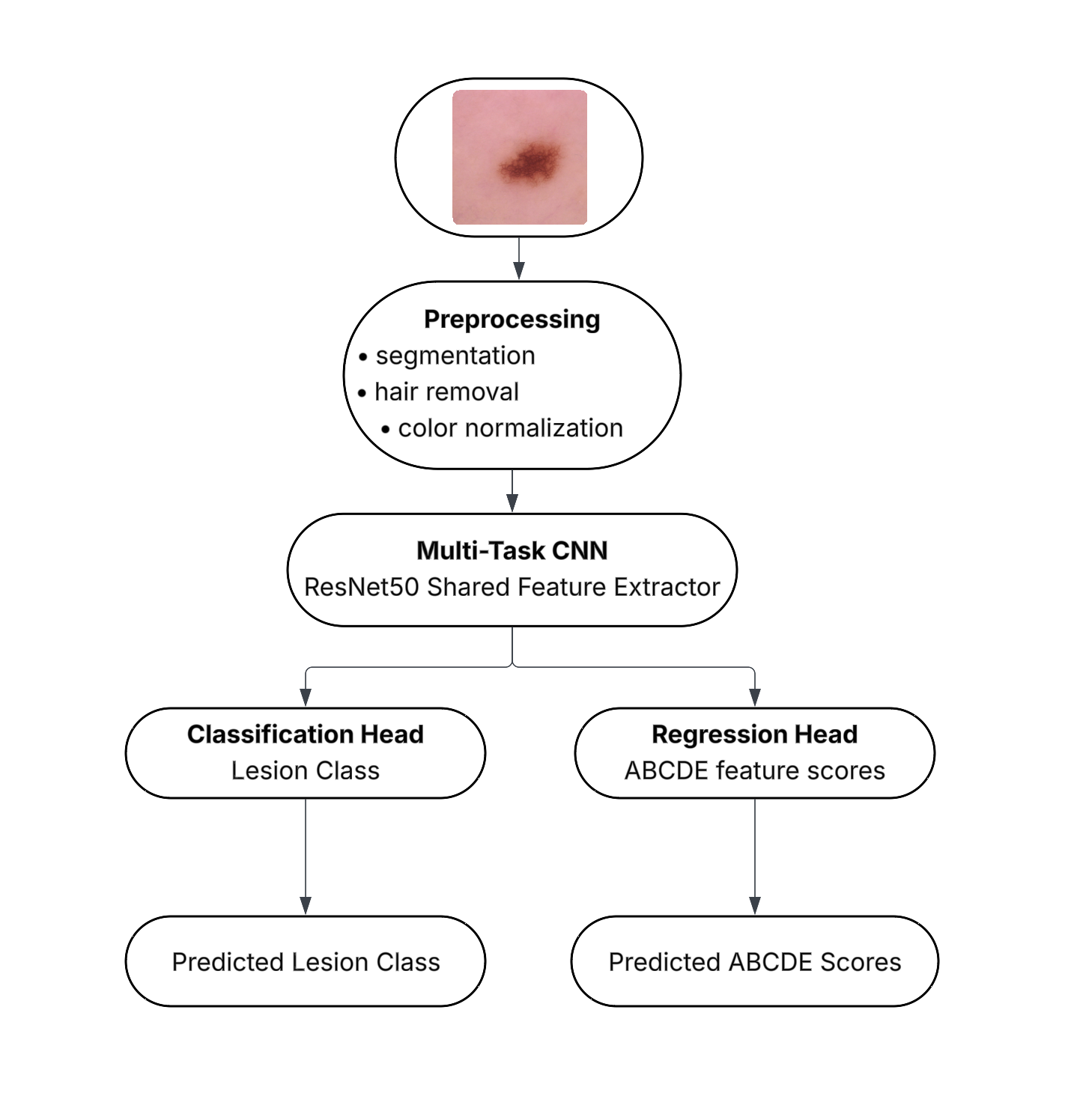}
    \caption{Overview of Multi-Task CNN framework for skin lesion classification and ABCDE feature prediction}
    \label{fig:MAEperclass}
\end{figure}

\newpage
\subsection{ABCDE Feature Engineering}

The ability to train this regression head needs ground truth numeric labels for the ABCDE features on each training image. Because the HAM10000 dataset does not provide these annotations, we derive approximate ABCD labels using image processing techniques \cite{Mader2018SkinCancerMNIST}. The main goal is to compute for each lesion image a set of feature values on a 0 - 1 scale that shows how strongly that criterion is present. These definitions are adopted from prior research on dermatology. 
\begin{itemize}

\item Asymmetry (A): We compare the lesion’s shape and color distribution across the axes for asymmetry \cite{Jutte2024Integrating}. First and foremost, we have to segment the lesion from the background skin. We then find the lesion’s major axis and minor axis (perpendicular to the major). After that, the lesion’s pixel mask is reflected across each axis, and we have to compute the overlap with the original. To illustrate, the score for asymmetry is 0 if it is perfectly asymmetric, and this value increases as symmetry decreases. Moreover, to consider color distribution, a structure similarity index or SSIM is computed between the two halves in the image intensity values \cite{Jutte2024Integrating}. The final asymmetry score is the average of the shape’s asymmetry and the color asymmetry metrics. In practice, round or evenly colored lesions yield a score near 0, but irregularly shaped or unevenly colored lesions yield a score closer to 1. The final value is stored in the variable “A”. 

\item Border Irregularity (B):  An irregular border is one that is ragged, notched, or blurred. We capture two aspects which are the shape irregularity and the sharpness of the border \cite{Jutte2024Integrating}. For shape irregularity, we compute the lesion’s convex hull and compare it to the actual border \cite{Jutte2024Integrating}. After that, we define the ratio between the lesion and the convex hull. A perfectly smooth-edged lesion has a ratio of 1.0, but lesions with indented edges or notches have a ratio less than 1. Higher values mean more irregularity. In order to analyze the border clarity, we take an approach similar to that of Jütte et al.; we compute the gradient magnitude along the lesion perimeter. We take the dermoscopic image, calculate the gradient around the border, and average it over all border pixels. Sharp borders have a higher gradient while fuzzy borders have a low gradient. We normalize this gradient value to [0,1]. For the overall border irregularity score, we combine shape and gradient: 

\[
B = 0.5 \cdot B_{\text{shape}} + 0.5 \cdot \bigl(1 - B_{\text{grad}}\bigr)
\]
In this equation, the gradient B value is inverted because a low gradient (fuzzy border) should increase the irregularity score. Thus, a lesion with a very jagged shape or a very blurred edge will have B near 1.

\item Color Variation (C): We measure how many different colors and shades are in the lesion. Common criteria for skin lesion images include colors like light brown, dark brown, black, blue-gray, white, and red \cite{Jutte2024Integrating}. A melanoma often has different colors. To quantify this value, we compute the dispersion of colors in the lesion. Specifically, we apply a clustering in color space to the lesion pixels \cite{Jutte2024Integrating}. The number of color clusters is determined, and then the dispersion index is calculated; this is the standard deviation of color distances of pixels from their cluster centroids as they are weighted by cluster size. This gives a value between 0 and 1 as 0 signifies a very unified color while 1 gives off a heterogeneity of colors. Additionally, we count the number of distinct color clusters. If there are more clusters, there is more color variety. We map the number of clusters to a 0–1 range as well. Our final color variation score C is a combination (we used the dispersion index primarily, and added a small increment for each distinct color beyond one). Benign single-color nevi typically score a value of C less than 0.2 while many melanomas with multiple shades of brown, black, and red would score more than 0.7.

\item Diameter (D): For the most part, lesions with a diameter greater than 6 millimeters are deemed as suspicious. However, these images lack a consistent physical scale as the zoom level varies \cite{Choi2024ABCEnsemble}. HAM10000 images come from different devices and magnifications \cite{Mader2018SkinCancerMNIST}. Unfortunately, no data on mm-per-pixel is given. That is why we will have to approximate the diameter relatively. After segmentation, we compute the maximum distance across the lesion (in pixels); this is the largest axis of the bounding box. We also compute the area-equivalent diameter, which is the diameter of a circle with the same area as the lesion. This max span is more robust for irregularly shaped lesions. To map this to a score, we make an assumption that an average nevus in the dataset (say 30 pixels across in the image) corresponds to about 5 mm; this is based on typical dermoscope resolution. 
\[
\mathbf{D} = \operatorname{clamp}\!\left(\frac{\text{max diameter (pixels)}}{p_{6\,\text{mm}}},\,0,\,1\right)
\]

This is where $p_{6\,\text{mm}}$ is the approximate pixel length of 6 millimeters. For better calibration, one could include a reference scale or use the dataset’s metadata; some ISIC images have ruler markers, but this is not the case in HAM10000. In our training labels, we set p6mm such that the top 5\% largest lesions get D close to 1.0 and small lesions get D near 0.2. The diameter values are thus relative. We treat D as a continuous variable because it influences risk in a graded way even though it is non-linear.

\item Evolving (E): Because the dataset does not contain time-series images, we could not directly measure or predict lesion evolution. That is why in this work we focus only on the static ABCD features for regression. The E criterion was not included in the training or loss function. Future work will investigate modeling lesion evolution. It will utilize time-series data to better capture this aspect.
\end{itemize}
It is still good to note that the automatically generated labels for A, B, C, and D are approximate. This is because errors in lesion segmentation or variations in image conditions can introduce noise. For example, if a lesion image is very close-up, the diameter estimate may falsely appear large; if lighting causes part of the border to fade into the skin, the border gradient metric might be low even if the border is fairly smooth. These issues could be mitigated with preprocessing. Applying a light hair removal filter (using inpainting for dark hairs) and a color normalization so that background skin has a consistent reference color across images. This can improve the accuracy of the feature computations. 

These computations output for each training image an A, B, C, and D value. We use these as the target outputs for the regression head. In this work, E is not directly mentioned but only represents the evolution or progression of the other A, B, C, and D features. 
\subsection{Model Training Strategy}

The model is trained on the HAM10000 dataset \cite{Tschandl2018HAM10000}; this dataset contains 10,015 dermoscopic images of pigmented lesions across 7 diagnostic categories. However, images are not evenly distributed across diagnostic categories (benign and melanoma). To handle this, we perform data balancing. Specifically, we use a combination of oversampling and class-balanced loss. During each training epoch, we sample images such that each class is roughly equally represented; this is done with the help of random oversampling of minority classes. We also weight the classification loss inversely proportional to class frequency. The data is split in this manner: 70\% of the images for training, 10\% for validation, and 20\% for testing. We also improve the training images with random horizontal or vertical flips, small rotations, zoom-in and out, and lighting adjustments. This helps to expose the network to varied appearances and also slightly simulates changes. 

In order to preprocess each image, we resize it to 224x224, apply the hair removal filter, normalize the color, and segment the lesions. For the loss functions, we have to combine the losses for both the heads of the model (classification and regression). The classification loss is the cross-entropy loss for the class prediction while the regression loss is the mean squared error between predicted and target scores for features A through D. We control the balance between these two losses using a weight parameter. In the main experiments, we set this value to be 1. However, it is also good to run other tests by setting that value to either 0 to train only classification or setting the value to a larger number than 1 to focus only on regression. This is very important because it shows why combining both tasks in this deep learning framework is beneficial.

For the optimization functions, the Adam optimizer is used with an initial learning rate of 0.0001. If the validation loss stops improving, we reduce the learning rate by a factor of 10. The training itself runs for around 100 epochs. To help prevent overfitting, we apply L2 weight decay with a value of 0.00001. All of these experiments use PyTorch and are trained on an NVIDIA A100 GPU.

For lesion classification, we evaluate standard metrics including the overall accuracy, per-class accuracy, precision, sensitivity, and F1-score for each class, and especially the melanoma detection specificity. Missing a melanoma would be the worst error. We also compute the balanced multi-class accuracy and the area under the ROC curve for the binary melanoma vs others task. For the ABCD feature regression, we measure the Pearson correlation between the predicted and ground-truth feature values on the test set; we also take a look at the mean absolute error. High correlation would mean that the model has learned to predict the features in relative terms even if there is some bias. Because the ground truth features (ABCD) are approximate, we focus more on correlation and rank ordering than exact error magnitude.We do not have ground truth for “E” on static images, so we cannot directly evaluate E predictions in the standard test set. To take care of the E, we will evaluate E in the context of simulated sequences. 

\subsection{Lesion Evolution Simulation}

Additionally, we propose a method to simulate how a lesion’s image and thereby, its ABCD features might change over time. This module operates in two possible modes. 

The first possible method for this is to use a generative adversarial network to generate a future version of a lesion image. Similar to the study presented by Jutte et al., a CycleGAN architecture is optimal here in order to translate an image from the domain of benign-appearing lesions to the domain of malignant-appearing lesions \cite{Jutte2024Integrating}. The CycleGAN learns to produce an image that does indeed retain the structure of the input, like the particular lesion’s shape and significant features, but it changes its appearance to resemble a melanoma. Evidently, according to what was mentioned above with the ABCD rule, higher asymmetry, more colors, and more irregular borders with the same background and context would show exactly that of a benign lesion evolving into a malignant melanoma. We also train the reverse mapping from a melanoma to a nevus using CycleGAN’s cycle-consistency; this is so that the network doesn’t simply always output a generic melanoma. To simulate a gradual evolution, we use frame interpolation \cite{Jutte2024Integrating}; this produces intermediate images that represent smooth transitions between benign and malignant appearances. Each frame is then analyzed by the multi-task CNN as it outputs ABCDE scores. By plotting the scores over frames, this yields a curve for each feature in order to exhibit the trend. These intermediate frames are not strictly physically accurate predictions because melanoma growth is not necessarily linear, but they are a good visual to provide clarity. The sequence can be thought of as a what-if scenario for how the lesion could evolve if it were to become malignant \cite{Jutte2024Integrating}. 

An alternate approach to this would be directly observing the feature values. The idea is to use the internal representation of the lesion, say the 2048-d feature from the CNN, and model how it would drift as the lesion evolves. The multi-task CNN’s feature extractor would be a state encoder in this scenario. Then, the “time steps” would be simulated by a small network that adjusts this state in the direction of malignancy. We train a simple feed-forward network to model how lesion features change over time. At each time step, it updates the latent feature vector zt by predicting how it will change. 

This is trained with the help of feature sequences extracted from images generated by the CycleGAN model. For each simulated sequence, we collect CNN features from each frame, and this gives us starting and ending feature pairs. The network learns to predict the difference between the initial and final features, and this helps it learn how features evolve as a lesion becomes more malignant. After training, we can simulate progression by applying small steps in the predicted direction. For example, starting with a benign lesion’s features, we can apply several steps and watch the ABCD scores increase; this parallels malignant transformation. While this method does not generate images, it is faster and does not need to store or run the image generator during inference. The model itself learns how features move towards malignant characteristics over time.

In the final system, the image generation approach above would be used primarily for visualization while the second approach would be used to verify that the direction of change in feature space corresponds to increasing ABCDE scores. Something that is important to note is a smooth change in segmentation masks for reliable ABCDE measurement; this is pointed out by Jütte et al. \cite{Jutte2024Integrating}. These sequences are not used to update the CNN. There should be no further training as that would require known ground truth of future states. They are purely for evaluation and demonstration. For this study, the latter approach (directly observing feature values) is performed to visualize plausible ABCD trajectories, and no image synthesis is used in the reported experiments.

\section{Results and Discussion}

On the HAM10000 dataset, the multitask CNN model did end up showing a strong classification performance. The overall accuracy was 89\%; it correctly classified 89\% of all test samples. Also, it had a balanced accuracy of 76\%. Some lesion types are much more common than others in the dataset, so the balanced accuracy gives equal weight to each class. This is regardless of how frequent or rare it is. It did end up performing well across all lesion types, even the less common ones.

\begin{figure}[h]
    \centering
    \includegraphics[width=0.5\linewidth]{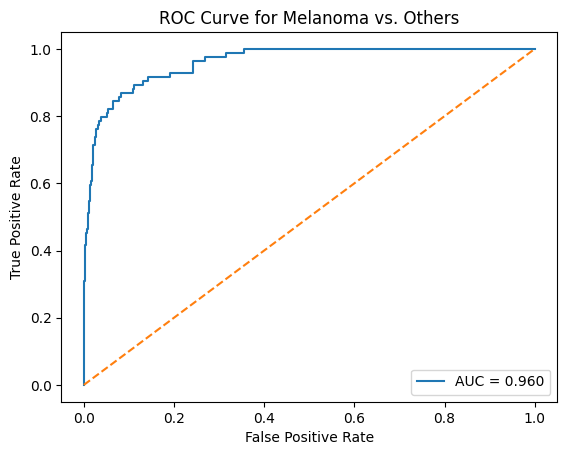}
    \caption{ROC Curve for Melanoma vs. Other lesions}
    \label{fig:roccurve}
\end{figure}

In the specific task of finding the difference between melanoma and all other lesion types, the model was effective. The ROC curve in Figure \ref{fig:roccurve} performed very well. The area under the curve (AUC) is 0.96. That means that the model is very efficient in classifying and at identifying melanomas correctly while minimizing false positives. 

\begin{figure}[h]
    \centering
    \includegraphics[width=0.7\linewidth]{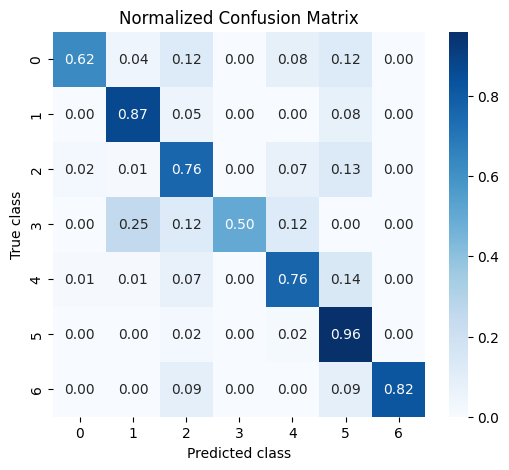}
    \caption{Normalized Confusion Matrix}
    \label{fig:NormalizedConfusionMatrix}
\end{figure}

Next, the confusion matrix in Figure \ref{fig:NormalizedConfusionMatrix} shows how often each lesion type was correctly or incorrectly predicted. The common classes like melanocytic nevus (nv) and basal cell carcinoma (bcc) models achieved very high recall; this means that it rarely missed them. However, for rarer classes such as dermatofibroma (df) and vascular lesions (vasc), the recall was lower, being around 50\% and 82\%. The model struggled more to correctly identify these. The recall for melanoma was about 76\%. Most of the errors happened between melanoma and visually similar benign lesions, like benign keratosis or nevus. This identifies the challenge of trying to look at the difference between classes that look very similar in dermoscopic images.
\newpage

The model overall had a very mixed accuracy regarding the predictions for the ABCD features. For the A, C, and D features, the model performs well. The average error is low, and the predictions are strongly correlated with the true values. This means the model is picking up on real, clinically meaningful patterns. However, the border irregularity was more difficult to find. Its prediction error is higher, and the model doesn’t show a strong relationship between predicted and true B scores. This could be because of noisy labels or the way border irregularity was measured. As it is broken down by class, B remains harder to predict across most lesion types. A, C, and D errors vary more by class. For example, melanoma cases have relatively accurate color predictions, but vascular lesions have larger diameter errors. 
\newpage
\begin{figure}[h]
    \centering
    \includegraphics[width=0.7\linewidth]{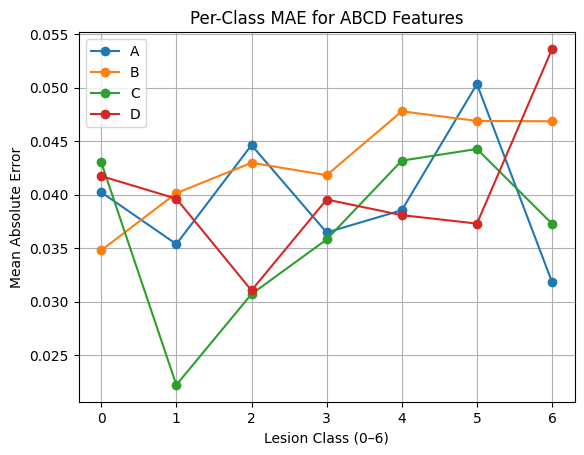}
    \caption{Per-Class MAE for ABCD Features}
    \label{fig:perclassmae}
\end{figure}

\begin{figure}[h]
    \centering
    \includegraphics[width=0.7\linewidth]{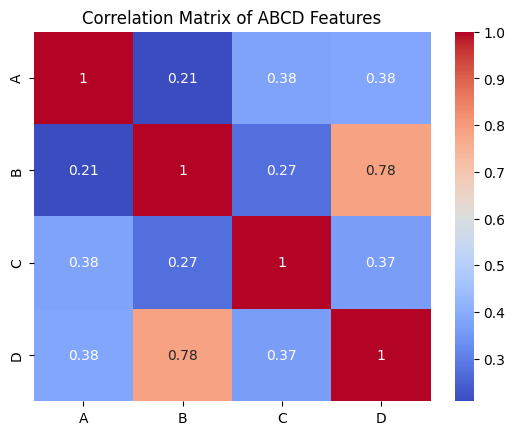}
    \caption{Correlation Matrix of ABCD features}
    \label{fig:corrmatrix}
\end{figure}

The correlation matrix in Figure \ref{fig:corrmatrix} shows how closely related the predicted ABCD features are to each other across all test samples. Each number represents the strength of the relationship between two features. Values closer to 1 mean that there is a stronger positive correlation.We observe a particularly strong correlation between border irregularity and diameter with a value of 0.78. This does make sense in a logical sense because larger lesions are more likely to have irregular borders. There are also moderate correlations between other features. Asymmetry shows a moderate relationship with color variation and diameter as they both are around 0.38. Even though ABCD features are mostly different, some are meaningfully related. This is how different visual traits of skin lesions often co-occur. Importantly, the features are not so highly correlated that they are redundant. This supports that the model can predict a variety of useful and separate clinical features rather than simply repeating the same signal.
\newpage

\begin{figure}[h]
    \centering
    \includegraphics[width=0.7\linewidth]{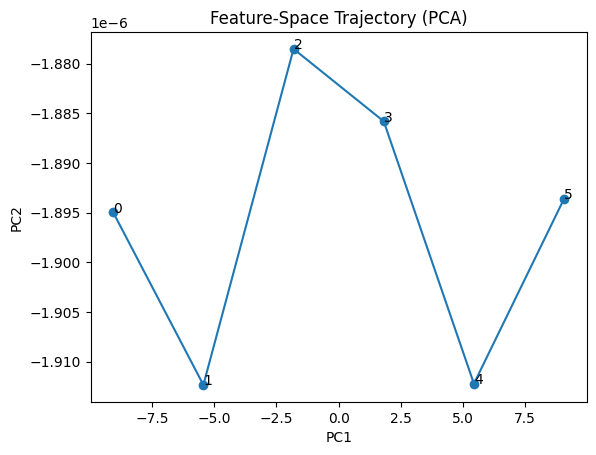}
    \caption{Feature-Space Trajectory (PCA)}
    \label{fig:pca}
\end{figure}

In addition, the plot below (Figure \ref{fig:pca}) shows how the lesion representation evolves in the model’s internal feature space when we simulate progression using latent-space interpolation. Each point (0 to 5) represents a step in the simulated evolution from the original lesion (step 0) toward a more malignant version (step 5). The axes PC1 and PC2 are the top two principal components from PCA. These are able to reduce the high-dimensional feature vectors into a 2D space while keeping most of the variation. 

The overall pattern shows a smooth and continuous trajectory. This is as the lesion steadily moves through the feature space. This shows that the model captures a consistent and directional change. This means that it has learned an internal sense of how lesions might evolve over time, even though it was not explicitly trained on temporal data. The zigzagging motion on PC2 shows how there would be subtle nonlinear shifts in features like color or shape. There is a more monotonic progression along PC1; this reflects a gradual move toward higher malignancy. This supports the idea that the model’s latent space encodes a clinically meaningful concept of lesion progression.
\newpage

\begin{figure}[h]
    \centering
    \includegraphics[width=0.7\linewidth]{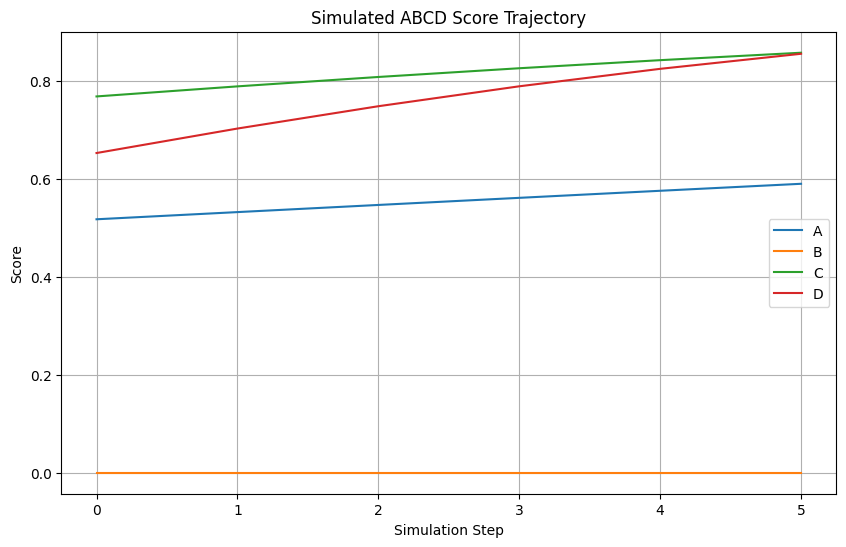}
    \caption{Simulated ABCD Score Trajectory}
    \label{fig:trajectory2}
\end{figure}

The plot below (Figure \ref{fig:trajectory2}) shows how the ABCD scores evolve across six simulated steps in the model’s learned feature space. Each line corresponds to one of the ABCD criteria. These scores are predicted by the model’s regression head as it interpolates from an original lesion in step 0 toward a hypothetical more malignant version in step 5. 

Visibly in the plot, A, C, and D all increase smoothly across the steps. This does indeed align with clinical expectations. As a lesion becomes more malignant, it typically becomes more asymmetric, shows greater variation in color, and increases in size. The upward trends in these scores suggest that the model captures these expected patterns of malignant progression; once again, it is important to note that it was not explicitly trained with temporal lesion data. In contrast, the B score of border irregularity was completely flat near zero. This confirms previous findings that the model struggles to predict this feature accurately. This is most likely due to noisy or insufficient training labels for border irregularity. The lack of progression in B shows a current limitation in modeling that particular clinical feature. 
\newpage

The results show that the multi-task CNN performs well overall as it combines strong lesion classification with interpretable ABCD feature predictions. It performs fairly well on key clinical tasks. For instance, with melanoma detection, it has an AUC of 0.96. It also learns to predict asymmetry, color variation, and diameter with strong accuracy. The model demonstrates that these features are not only predictable but also embedded meaningfully within the network’s latent space. This is shown by the smooth trajectory and increasing malignancy indicators in the evolution simulation. 

One clear limitation is the poor performance in predicting border irregularity (B). This likely stems from how B was labeled. It was most likely based on simple segmentation heuristics rather than clinical assessment. This introduced noise and weakened both regression and simulated trends. Also, the dataset’s class imbalance affected both classification and regression accuracy for rare lesion types. Another limitation is that the evolution simulation was performed in latent feature space and not directly on images. This visual progression remains abstract.

This system could assist clinicians not only in diagnosing skin lesions but also in interpreting why the model made a decision. This is through the ABCD feature outputs. The evolution simulation can offer “what-if” previews of how lesions might progress toward malignancy, and this can support patient education and monitoring.

To improve border irregularity prediction, better labeling methods, such as dermatologists or advanced segmentation techniques are needed. Also, making sure that the classes are more balanced with augmentation or resampling could improve fairness across the lesion types. A good next step would be to generate simulated lesion images. It would use a GAN-based method to complement the feature-based evolution, and this would offer more intuitive visual feedback. Lastly, it would be good to incorporate expert-reviewed or longitudinal data that would allow supervised training of the “E” component and enhance clinical realism.

\section{Conclusion}

To summarize, we developed a deep learning model to accurately classify skin lesions, explain its predictions with the help of ABCD features, and simulate how a lesion might change over time. The model provides robust quantification of asymmetry, color, and diameter, but struggles with border irregularity. Still, it offers a clear and interpretable way to analyze lesions and track potential progression. In the future, we aim to improve the border score, add realistic image evolution, use expert-labeled data for the “E” feature, and test the model on more diverse datasets.



\bibliographystyle{apalike}

\bibliography{sample}

\end{document}